\documentclass[10pt,twocolumn,letterpaper]{article}

\usepackage[pagenumbers]{cvpr} 

\usepackage{graphicx}
\usepackage{amsmath}
\usepackage{amssymb}
\usepackage{booktabs}
\usepackage{pifont}
\usepackage{appendix}
 \usepackage{array,multirow}
 \usepackage{float}

\usepackage{microtype}
\usepackage{booktabs} 
\usepackage{multirow}
\usepackage{caption}
\usepackage{bbm}
\usepackage{setspace}
\usepackage[ruled,vlined]{algorithm2e}
\usepackage{xcolor}

\def\eg{\emph{e.g}.} 
\def\ie{\emph{i.e}.} 
 
\def\etc{\emph{etc}.}

\usepackage[pagebackref,breaklinks,colorlinks]{hyperref}

\usepackage[capitalize]{cleveref}
\crefname{section}{Sec.}{Secs.}
\Crefname{section}{Section}{Sections}
\Crefname{table}{Table}{Tables}
\crefname{table}{Tab.}{Tabs.}

\begin{document}

\title{Beyond RGB: Scene-Property Synthesis with Neural Radiance Fields}
\author{
{Mingtong Zhang\thanks{Equal contribution} $^{ ,1}$ \quad Shuhong Zheng\footnotemark[1] $^{ ,1}$ \quad Zhipeng Bao$^{2}$ \quad Martial Hebert$^{2}$ \quad Yu-Xiong Wang$^{1}$} \\
{ $^1$University of Illinois Urbana-Champaign \qquad $^2$Carnegie Mellon University } \\
  \texttt{\small  \{mz62, szheng36, yxw\}@illinois.edu \qquad \{zbao, hebert\}@cs.cmu.edu } \\
}
\maketitle

\begin{abstract}
 Comprehensive 3D scene understanding, both geometrically and semantically, is important for real-world applications such as robot perception. Most of the existing work has focused on developing data-driven discriminative models for scene understanding. This paper provides a new approach to scene understanding, from a synthesis model perspective, by leveraging the recent progress on implicit 3D representation and neural rendering. Building upon the great success of Neural Radiance Fields (NeRFs), we introduce Scene-Property Synthesis with NeRF (SS-NeRF) that is able to not only render photo-realistic RGB images from novel viewpoints, but also render various accurate scene properties (e.g., appearance, geometry, and semantics). By doing so, we facilitate addressing a variety of scene understanding tasks under a unified framework, including semantic segmentation, surface normal estimation, reshading, keypoint detection, and edge detection. Our SS-NeRF framework can be a powerful tool for bridging generative learning and discriminative learning, and thus be beneficial to the investigation of a wide range of interesting problems, such as studying task relationships within a synthesis paradigm, transferring knowledge to novel tasks, facilitating downstream discriminative tasks as ways of data augmentation, and serving as auto-labeller for data creation.
\end{abstract}

\section{Introduction}
\label{sec:intro}
Consider a domestic robot that is navigating in a room and performing various types of household tasks. To do so, the robot needs a comprehensive geometric and semantic understanding of the scene, uncovering the complete 3D spatial layout, functional attributes, semantic labels of the scene, its constituent objects, \etc~\cite{naseer2018indoor}. Most of the existing work on 3D scene understanding has focused on developing data-driven {\em discriminative} models for various scene analysis problems~\cite{He_2019_ICCV,joseph2021towards}, such as semantic segmentation, object detection, and surface normal estimation. By contrast, this paper introduces a novel perspective for scene understanding -- instead of developing discriminative models for different scene understanding tasks, we synthesize various paired scene properties by leveraging the underlying 3D scene representation and neural rendering.

An important first step toward synthesizing various scene properties is rendering photo-realistic images. One of the most influential advances in this direction is Neural Radiance Fields (NeRF)~\cite{mildenhall2020nerf} which, given a handful of images of a static scene, learns an implicit volumetric representation of the scene that can be rendered from novel viewpoints. By sampling the coordinates along each camera ray from various views, NeRF represents a complex scene as a continuous 5D implicit function with a multilayer perceptron network, which regresses from a single 5D coordinate to a single volume density and view-dependent RGB color. In the end, NeRF accumulates those colors and densities into a 2D image through volume rendering. The implicit representation is optimized by minimizing the residual between synthesized images and ground-truth from various views. NeRF has inspired significant follow-up work that has primarily focused on improving the quality of rendered images~\cite{schwarz2020graf,niemeyer2021giraffe}, speeding up the training and rendering time~\cite{lindell2021autoint,reiser2021kilonerf,garbin2021fastnerf,deng2021depth}, handling unbounded scenes~\cite{kaizhang2020} and dynamic scenes~\cite{martinbrualla2020nerfw}, and incorporating uncertainty~\cite{shen2021stochastic}.

\begin{figure}[t]
		\centering
        \includegraphics[width = 0.9 \linewidth]{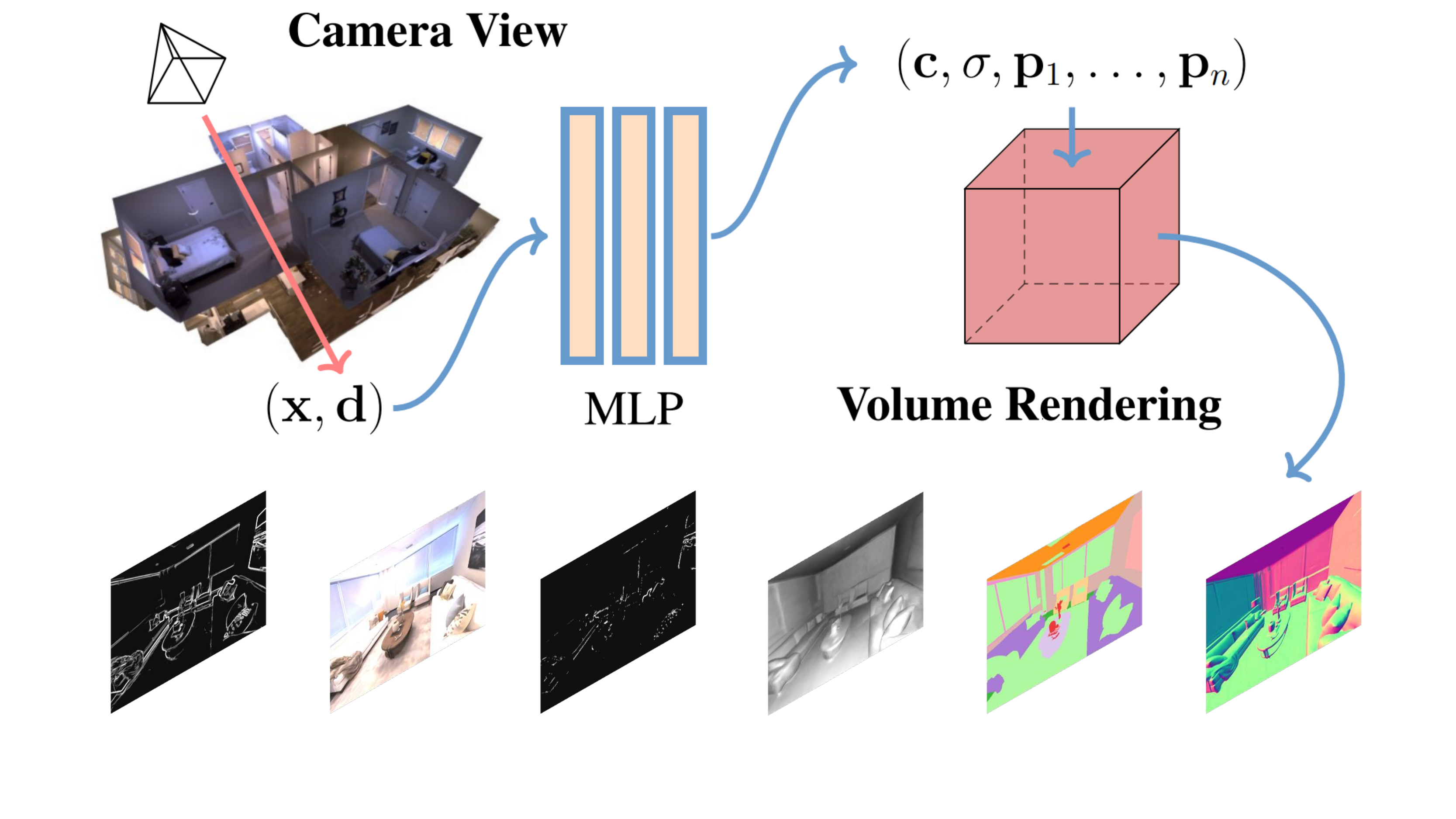}	
		\vspace{-15 pt}
		\caption{We represent the scene as an implicit function and develop a {\em versatile} neural scene representation, SS-NeRF, that is able to not only render images from novel viewpoints, but also render various scene properties (\eg, appearance, geometry, and semantics) paired with the synthesized images, under a unified framework.
		}
		\label{fig:teaser}
		\vspace{-15 pt}
\end{figure}

In this paper, we are interested in a different question: {\em Could this implicit representation be extended to synthesize richer scene properties beyond RGB color?} The answer is {\bf yes}. As illustrated in Fig.~\ref{fig:teaser}, we develop a NeRF-style model that is able to render not only photo-realistic RGB images from novel viewpoints, but also various accurate scene properties corresponding to the synthesized images, {\em under a unified framework}. This thus facilitates comprehensive scene understanding including semantic segmentation, surface normal estimation, reshading, keypoint detection, and edge detection. We call our framework  {\em Scene-Property Synthesis with NeRF} (SS-NeRF).

Naturally, we find some of the scene properties are sensitive to the observation directions, while others are not (\eg, semantic labels), for which the viewing direction input of the original NeRF model is redundant. Therefore, we adopt two branches to take care of these different properties (shown in Fig.~\ref{fig:model}) that consider or ignore the view direction input ($\theta, \phi$) respectively, so that the proposed SS-NeRF model is able to deal with different types of properties in a coherent way, yielding realistic synthesis for all of them. Moreover, the learned scene representation is shareable and beneficial across different properties, leading to the result that SS-NeRF is able to generalize from synthesizing a single property to multiple properties. 

As a general, flexible framework, SS-NeRF further facilitates the investigation of a variety of interesting problems.
For example, within the SS-NeRF framework, we analyze the relationship among different scene properties through both multi-task learning and knowledge transfer. We show that a learned implicit geometric and semantic representation enables the flow of knowledge across different synthesis tasks, so that they can benefit one another. While similar phenomena have been widely investigated in the regime of discriminative models such as Taskonomy~\cite{zamir2018taskonomy}, they are largely under-explored within a synthesis model. Moreover, we explore two applications of SS-NeRF. We show that the examples synthesized by SS-NeRF (RGB images paired with scene properties) can be used effectively as augmented data for improving the corresponding downstream discriminative tasks. In addition, we show that, because of its learned underlying 3D geometry and scene representations,  SS-NeRF can work as an auto-labeller to refine the pseudo-labels produced by state-of-the-art discriminative models.  

{\bf Our contributions} are four-fold: {\bf (1)} We propose a novel solution SS-NeRF to scene understanding from the perspective of learning a synthesis model. To the best of our knowledge, SS-NeRF is {\em the first work} that extends NeRF to simultaneously rendering photo-realistic novel-view images and {\em various} corresponding scene properties.
{\bf (2)} We instantiate SS-NeRF with five popular scene properties, including semantic labels, surface normal, shading, keypoints, and edges. Intriguingly, as a {\em versatile neural scene representation}, SS-NeRF is shown to outperform a hybrid strategy that trains NeRF (for rendering images) and task-specific discriminative models (for predicting scene proprieties) separately.
{\bf (3)} We show that our SS-NeRF framework is a powerful tool for {\em bridging generative learning and discriminative learning}, bringing new insight into the investigation of relationships between different properties and tasks via multi-task learning or knowledge transfer within a synthesis paradigm.
{\bf (4)} We further demonstrate that SS-NeRF can be beneficial to a variety of problems, such as facilitating downstream tasks as ways of data augmentation and serving as auto-labeller for data creation.

\section{Related Work}
\label{sec:related}
{\bf Novel-View Synthesis} aims to generate a target image with an arbitrary camera pose from one or few given source images~\cite{tucker2020single}. Generative Adversarial Networks~\cite{NIPS2014_5ca3e9b1} (GANs)-based models have shown promising results for synthesizing photo-realistic images of novel views~\cite{Karras_2020_CVPR,zhao2018modular,Goetschalckx_2019_ICCV,chen2016infogan,nguyen2019hologan,bao2020bowtie}. Though some work also investigates explicitly modeling geometrical properties~\cite{greff2019multi,burgess2019monet} or introducing 3D shape representations as inductive bias~\cite{Zhu-2018-125683,wu2016learning,henderson2020leveraging}, these models still cannot learn implicit 3D representations.

\textbf{Implicit Scene Representation} encodes scenes into feature vectors for novel-view synthesis. Combining the implicit neural model and the volume rendering technology, Neural Radiance Field (NeRF) \cite{mildenhall2020nerf} achieves impressive performance in novel-view synthesis of complicated scenes. It learns an implicit 3D geometry representation of scenes with perceptron networks and synthesizes views by querying along camera rays with classic volume rendering techniques. Some following work further improves the generalization capability~\cite{schwarz2020graf,gu2021stylenerf,yu2021pixelnerf,attal2021t}, compositionality~\cite{niemeyer2021giraffe,kaizhang2020,Ost_2021_CVPR,guo2020object}, and efficiency of inference~\cite{lindell2021autoint,reiser2021kilonerf,garbin2021fastnerf,deng2021depth}. Inductive biases, such as depth and multi-view consistency, are also introduced to facilitate NeRF-style architectures~\cite{wei2021nerfingmvs,oechsle2021unisurf,wang2021neus}. Furthermore, the volume rendering technique and underlying 3D geometry and scene representation are also applied to benefit other model structures~\cite{meng2021gnerf,kohli2020semantic,sajjadi2021scene}.

While most of the NeRF-based work still focuses on the RGB synthesis task, some explorations have been made to extend NeRF from RGB synthesis to other scene properties. \cite{oechsle2021unisurf} learns the surface of the objects together with the color and density, leading to efficient and effective rendering. \cite{yariv2021volume} improves geometry representation and reconstruction in neural volume by modeling the surface density. Semantic-NeRF~\cite{Zhi:etal:ICCV2021} also extends the NeRF-style architecture to semantic annotations, which can be viewed as a {\em special instance} of our framework, and explores several valuable applications. Different from such work, SS-NeRF scales from RGB synthesis to other pixel-level scene properties, from individual to {\em multiple} properties, with a shared 3D geometry and scene representation. 

Recent methods for \textbf{Scene Understanding} have gained impressive performance in semantic segmentation~\cite{kundu2020virtual,paul2020domain,fan2020sne,He_2019_ICCV}, object detection~\cite{bolya2020tide,zhu2020soft,joseph2021towards}, 3D \& visual reasoning~\cite{park2019deepsdf,mescheder2019occupancy,Yang_2021_ICCV,baradel2018object,chen2018iterative,johnson2017inferring}, \etc~Despite great achievements, few of them focus on understanding scenes from a synthesis model perspective. In comparison, SS-NeRF considers an implicit representation of 3D shape and scene properties, allowing for knowledge transfer and feature sharing across different tasks and thus capturing the underlying image generation mechanism for more comprehensive scene understanding than being done within individual tasks.

\textbf{Multi-task Learning} aims to jointly solve different tasks through leveraging shared knowledge from related tasks~\cite{crawshaw2020multi}. Recent work mainly uses either soft parameter sharing~\cite{misra2016cross,yang2016trace} or hard parameter sharing~\cite{kokkinos2017ubernet,Doersch_2017_ICCV} strategies~\cite{ruder2017overview}. Beyond solving multi-task learning, the task relationships among different tasks have also been studied. \emph{Taskonomy} and the following work~\cite{zamir2018taskonomy,standley2019tasks,bao2021generative,sun2019adashare,zamir2020robust} extensively exploit the task relationships to gain the best performance. Compared with prior work, SS-NeRF, as a synthesis model, can also be scaled to solve multiple visual tasks jointly, and further investigate task relationships.

\section{Methodology}
\label{sec:method}

\begin{figure}[t]
\vspace{-15 pt}
		\centering
        \includegraphics[width = 0.85 \linewidth]{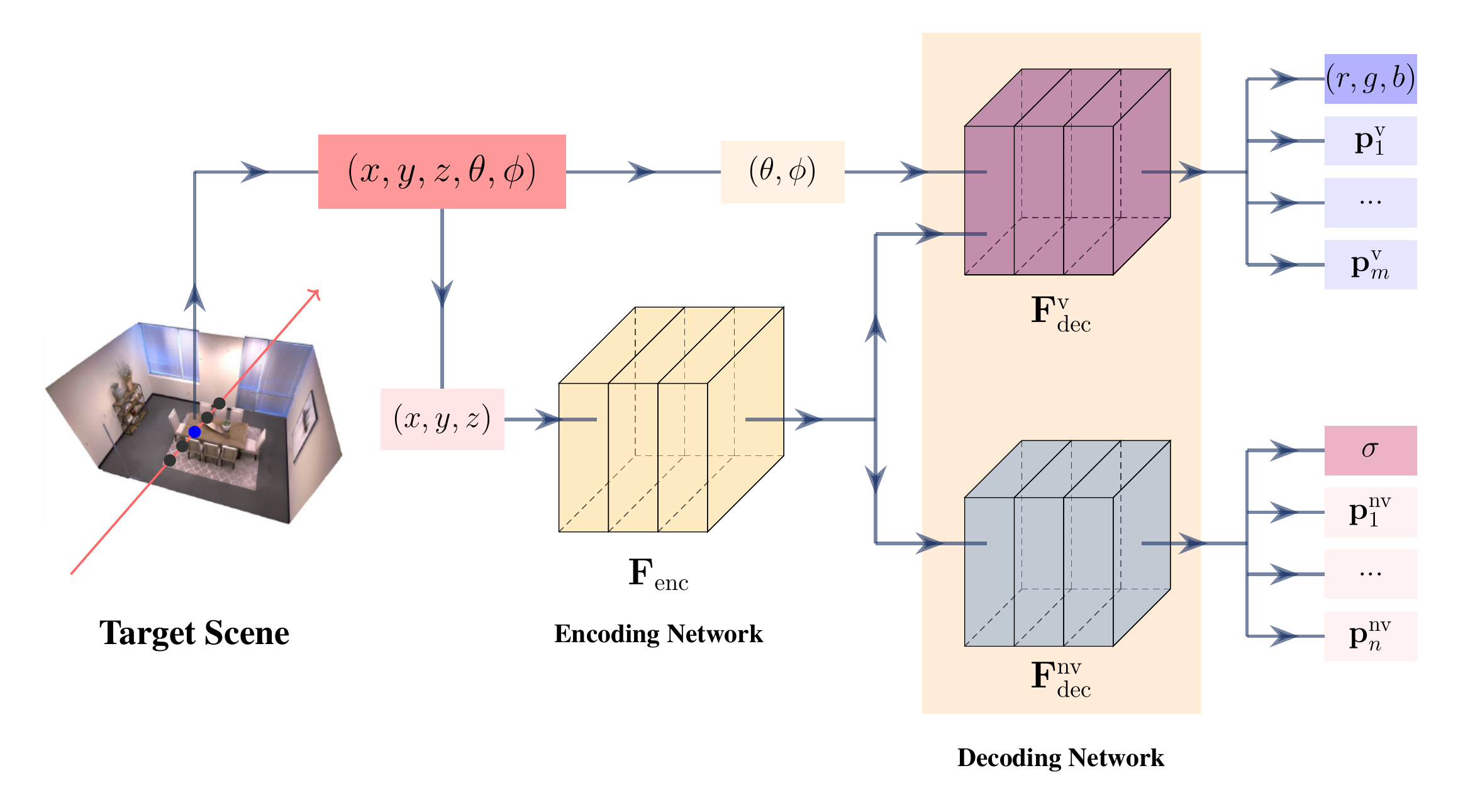}
        \vspace{-15 pt}
		\caption{SS-NeRF architecture. The model takes the 3D coordinates and view directions as input and is able to synthesize different paired scene properties. SS-NeRF uses a shared scene encoding network $\mathbf{F}_{\mathrm{enc}}$ to conduct the 3D positional embedding, followed by two separate decoding networks $\mathbf{F}_{\mathrm{dec}}^{\mathrm{v}}$ and $\mathbf{F}_{\mathrm{dec}}^{\mathrm{nv}}$ which produce scene property predictions. $\mathbf{F}_{\mathrm{dec}}^{\mathrm{v}}$ considers the view input, while $\mathbf{F}_{\mathrm{dec}}^{\mathrm{nv}}$ does not.
		}
		\label{fig:model}
		\vspace{-15 pt}
	\end{figure}

Fig.~\ref{fig:model} illustrates the architecture of our proposed SS-NeRF framework (Scene-Property Synthesis with NeRF). In this section, we first introduce the basic concept of Neural Radiance Fields, followed by the problem setting and innovation of SS-NeRF. Finally, we describe our SS-NeRF design in detail and instantiate it with five representative tasks in the context of scene understanding.

\subsection{Neural Radiance Fields}
\label{sec:method-NeRF}

Given a 3D point  $\mathbf x = (x, y, z)$ and a viewing direction $\mathbf d = (\theta, \phi)$, NeRF~\cite{mildenhall2020nerf} learns an implicit scene representation $f$ to map the 5D input to an RGB color $\mathbf c=(r, g, b)$ and volume density $\sigma$: $f(\mathbf{x}, \mathbf{d}) \mapsto (\mathbf{c}, \sigma)$. 

Then NeRF calculates the single pixel color value by tracking and sampling the camera ray $\mathbf{r}(t) = \mathbf{o} + t\mathbf{d}$, which is emitted from the center of the camera plane $\mathbf{o}$ in the direction $\mathbf{d}$. Specifically, it randomly samples $M$ quadrature points $\{t_m \}_{m=1}^M$ with color $\mathbf c(t_m)$ and density $\sigma(t_m)$ between the near boundary $t_n$ and far boundary $t_f$. Then the approximated color of that pixel is given by:
\begin{equation}
    \mathbf{\hat{C}}(\mathbf{r}) = \sum_{m=1}^M \hat{T}(t_m)\alpha (\delta_m\sigma(t_m))\mathbf{c}(t_m) ,
    \label{vr}
\end{equation}
where $\delta_m$ is the distance between the two consecutive sample points ($\delta_m =\| t_{m+1} - t_m \|$), $\alpha(d) = 1 - \exp(-d)$, and 
\begin{equation}
    \hat{T}(t_m) = \exp\left(- \sum_{j = 1}^{m-1} \delta_j \sigma(t_j)  \right)
    \label{tt}
\end{equation}
denotes the accumulated transmittance. 

\vspace{-5pt}
\subsection{Innovation and Problem Setting}
\label{sec:method-problem}
\vspace{-5pt}

\textbf{Innovation:}
NeRF learns implicit 3D geometry and scene representation with perceptron networks. Our key insight is that this kind of geometry-aware representation can be applicable to and useful for not only RGB color, but also other scene properties, since it is internally shareable. Moreover, such representation addresses limitations of both {\em discriminative} models (generalization to novel views) and {\em GAN-based generative} models (generalization from image synthesis to other tasks) coherently. It thus provides a novel synthesis perspective for scene understanding and introduces new potential for a wide range of applications.

\textbf{Problem Setting:}
We generalize the basic NeRF setting from single RGB synthesis to rendering additional {\em pixel-wise scene properties} (\eg, semantic labels, edges, surface normal, \etc). Specifically, for a certain scene property $P_i$, we aim to learn a function $f_i$ to estimate its values $\mathbf{p}_i$ for each 3D location and view direction: $f_i(\mathbf{x}, \mathbf{d}) \mapsto  \mathbf{p}_i$. 

Moreover, since the implicit function encodes the geometry, shape, and texture information of the scene, which are shareable across different property prediction tasks, we argue that different properties can be learned together with shared knowledge. Thus, we further formulate the ``Scene-Property Synthesis'' problem as follows: Given a collection of {\em K} scene properties $\mathcal{P} = \{P_k \}_{k=1}^K$, we aim to build a representation function $f$ that can map the 3D coordinates and the view directions to the corresponding property values $f(\mathbf{x}, \mathbf{d}) \mapsto \{ \mathbf{p}_k \}_{k=1}^K$.

\subsection{SS-NeRF}
\label{sec:method-instance}
\textbf{Model Architecture:}
To solve this novel problem, we propose SS-NeRF, whose model architecture is shown in Fig.~\ref{fig:model}. Note that, while in principle our framework is applicable to more powerful NeRF variants for improved performance, here we focus on the basic NeRF model~\cite{mildenhall2020nerf}, showing the effectiveness and generalizability of SS-NeRF {\em without other advanced components and design choices}. Concretely, the whole model learns to map the 5D vector (3D coordinates and 2D view directions) to the corresponding scene properties; then we render the scene property ``images'' with the volume rendering technique used by \cite{mildenhall2020nerf}. 

We first adopt a shared positional encoder $\mathbf{F}_\mathrm{enc}$ to build feature embeddings $e_\mathbf{x}$ for the 3D coordinates $(x,y,z)$: 
\begin{equation}
e_\mathbf{x} = \mathbf{F}_{\mathrm{enc}}(x,y,z).
\end{equation}
Some scene properties (\eg, semantic labels) are not sensitive to the viewing direction, so that the view input is redundant. Therefore, we adopt two types of decoding networks $\mathbf{F}_{\mathrm{dec}}^{\mathrm{v}}$ and $\mathbf{F}_{\mathrm{dec}}^{\mathrm{nv}}$ inspired by~\cite{Zhi:etal:ICCV2021}. The $\mathbf{F}_{\mathrm{dec}}^{\mathrm{v}}$ takes the additional view input $\mathbf{d} = (\theta, \phi)$ together with the encoded coordinates to make the predictions for property $P^v_i$, while $\mathbf{F}_{\mathrm{dec}}^{\mathrm{nv}}$ predicts scene property $P^{\mathrm{nv}}_j$ directly with the encoded coordinates:
\begin{equation}
\hat{\mathbf{p}}^{\mathrm{v}}_i  = \mathbf{F}_{\mathrm{dec}}^{\mathrm{v}}(e_\mathbf{x}, \theta, \phi); \quad \hat{\mathbf{p}}^{\mathrm{nv}}_j  = \mathbf{F}_{\mathrm{dec}}^{\mathrm{nv}}(e_\mathbf{x}).
\end{equation}
In practice, in our preliminary experiment, we tried these two modeling strategies for each scene property and adopted the one that works better in all the following experiments. We have also validated the necessity of this two-branch model design with ablations in Sec.~\ref{sec:ablation}. 

The simplest working scenario of SS-NeRF is to predict a single scene property. However, by adding more decoding branches, the proposed model is able to predict multiple properties, leading to the generalization from a single task to multiple tasks. In Sec.~\ref{sec:applications}, we discuss the application for multi-task learning with SS-NeRF. Notice that the density $\sigma$ is always required to do the volume rendering for either single property or multiple properties, and the color is the most informative scene property, we treat them as the fixed outputs for our SS-NeRF model and add other properties upon this basic model.

\noindent
\textbf{Instantiation and Optimization of SS-NeRF:}
We instantiate SS-NeRF with five representative scene properties that are important in practice~\cite{zamir2018taskonomy,standley2019tasks}, together with the color image synthesis. These properties are: {\bf Semantic Labels (SL), Surface Normal (SN), Shading (SH), Keypoints (KP), and Edges (ED)}. We adopt $\mathbf{F}_\mathrm{dec}^\mathrm{v}$ for SH, KP, and ED; and $\mathbf{F}_\mathrm{dec}^\mathrm{nv}$ for SL and SN.

During the optimization of SS-NeRF, we adopt the hierarchical volume sampling strategy proposed by~\cite{mildenhall2020nerf}. That is, we first randomly pick some ``coarse'' sample points and then produce a more informed sampling of ``fine'' points that are biased towards the relevant parts of the volume. We also use task-specific objectives for these different properties. For the color image synthesis, we adopt the mean square error (MSE):
\begin{equation}
\small 
    \mathcal{L}_\mathrm{rgb} = \mathcal{L}_\mathrm{MSE} =\sum_{\mathbf{r}\in\mathcal{R}}\left[\left\Vert \hat{\mathbf{p}}_c(\mathbf{r})-\mathbf{p}(\mathbf{r})\right\Vert ^2_2 + \left\Vert\hat{\mathbf{p}}_f(\mathbf{r})-\mathbf{p}(\mathbf{r})\right\Vert^2_2\right],
    \label{MSE}
\end{equation}
where $\mathbf{p}(\mathbf{r}), \hat{\mathbf{p}}_c(\mathbf{r}), \hat{\mathbf{p}}_f(\mathbf{r})$ are the ground-truth, coarse volume prediction, and fine volume prediction for property $P$, respectively. $\mathcal{R}$ is the set of rays $\mathbf{r}$ in each batch. The MSE loss is also used for the surface normal prediction. For semantic label prediction, we use the cross entropy loss function:
\begin{equation}
\small
\mathcal{L}_\mathrm{seg}=\sum_{\mathbf{r}\in \mathcal{R}}\left[\sum^L_{l=1}s^l(\mathbf{r})\log \hat s^l_c(\mathbf{r}) + \sum^L_{l=1}s^l(\mathbf{r})\log \hat s^l_f(\mathbf{r})\right],
\end{equation}
where $s^l, \hat s^l_c, \hat s^l_f$ are the ground-truth, coarse volume prediction, and fine volume prediction of multi-class semantic probability of class $l$, respectively. Coarse and fine predictions $\hat s^l_c, \hat s^l_f$ are processed by a softmax layer after volume rendering. For shading, keypoints, and edges, we adopt the $\mathcal{L}_1$ loss:
\begin{equation}
\small 
    \mathcal{L}_\mathrm{ABSE}=\sum_{\mathbf{r}\in\mathcal{R}}\left[\left\Vert \hat{ \mathbf{p}}_c(\mathbf{r})-\mathbf{p}(\mathbf{r})\right\Vert_1 + \left\Vert\hat{ \mathbf{p}}_f(\mathbf{r})-\mathbf{p}(\mathbf{r})\right\Vert_1\right].
    \label{L1_loss}
\end{equation}
The final loss is the weighted sum of photo-metric loss and the standard loss of the specific task as:
\begin{equation}
    \mathcal{L}_\mathrm{whole}=\mathcal{L}_\mathrm{rgb} + \sum_{P_i \in \mathcal{P}} \lambda_{P_i} \mathcal{L}_{P_i},
\end{equation}
where $\mathcal{P}=\{P_{SL}, P_{SN}, P_{SH}, P_{KP}, P_{ED}\}$ is the set of properties, and $\lambda_{P_i}$ is the corresponding weight. 

\noindent
\textbf{Modeling of Surface Normal:} Among all the five scene properties, the surface normal is a special one that is of a vector form, whose projection in the image depends on the camera pose. To better model this property, we use $\mathbf{F}_\mathrm{dec}^\mathrm{nv}$ as the decoding network but introduce an additional input of encoded camera pose to directly synthesize the encoded normal with the volume rendering technique. 

\section{Experimental Evaluation}
\label{sec:exp}

\begin{figure*}[t]
    \centering
    \includegraphics[width =  \linewidth]{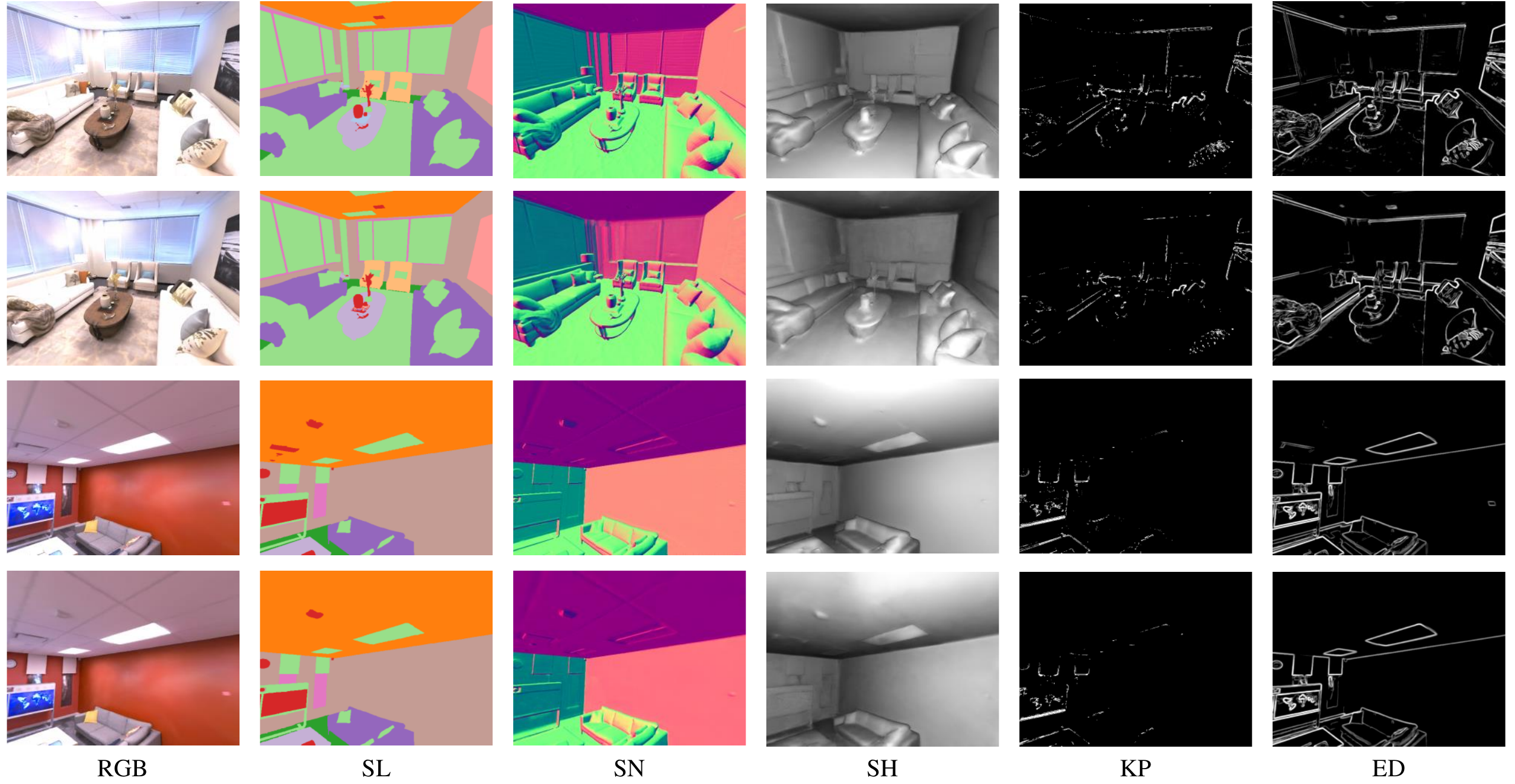}
    \vspace{-20 pt}
	\caption{Two qualitative results of testing views on Replica~\cite{straub2019replica}. {\bf Top row:} ground-truth; {\bf Bottom row:} our synthesis results. The synthesized RGB images are from the SL task. SS-NeRF is able to render {\em realistic and matched} RGB images and other properties.}
		\label{fig:main}
		\vspace{-15 pt}
\end{figure*}

In this section, we evaluate SS-NeRF. We start with the experimental setting in Sec.~\ref{sec:exp_setting}, followed by the quantitative and qualitative evaluation for all the five scene properties (Sec.~\ref{sec:withrgb}). In Sec.~\ref{sec:ablation}, we ablate the model performance without the color image synthesis branch and different decoding networks. We then make further explorations for SS-NeRF, including knowledge transfer and task relationships, data augmentation for downstream discriminative tasks, and also real-world auto-labeller applications (Sec.~\ref{sec:applications}). Finally, we discuss the limitations and future work in Sec.~\ref{sec:limitation}.

\subsection{Experimental Setting}
\label{sec:exp_setting}
\textbf{Datasets:} We first conduct extensive experiments on the commonly-used Replica~\cite{straub2019replica} dataset. The Replica dataset is a high-quality synthetic scene dataset containing photo-realistic 3D modelings for 18 scenes in total. Following~\cite{Zhi:etal:ICCV2021}, we conduct experiments on four scenes and each scene contains 50 frames at resolution $640 \times 480$. We have also validated the robustness of our model on the BlendedMVS dataset~\cite{yao2020blendedmvs} and investigated the application of SS-NeRF on complex real-world scenes (LLFF) collected by~\cite{mildenhall2020nerf} and \cite{mildenhall2019local}. The image resolutions of these two datasets are $768 \times 576$ and $4032 \times 3024$, respectively. We follow the same processing as NeRF~\cite{mildenhall2020nerf} for LLFF.

\textbf{Target Properties:} Following the observation in~\cite{standley2019tasks}, we include five important scene properties other than the RGB color in our experiments: Semantic Labels (SL), Surface Normal (SN), Shading (SH), Keypoint (KP), and Edge (ED). For Replica, we map the original 88-way semantic classes to the commonly-used NYUv2-13~\cite{Silberman:ECCV12,Eigen_2015_ICCV} format. 


\textbf{Scene Annotations:} We render the missing annotations by ourselves. The surface normal is derived from the depth by $SN(x,y,z) = (- \frac{dx}{dz}, -\frac{dy}{dz},1)$, where ($x,y,z$) are the 3D coordinates and $\frac{dx}{dz}$, $\frac{dy}{dz}$ are the gradients of z with respect to $x$ and $y$, respectively. Edges are rendered by a Canny~\cite{canny1986computational} detector; Keypoints are derived from SURF~\cite{bay2006surf}; Shadings are rendered by a pre-trained model, XTConsistency~\cite{zamir2020robust}. 

\textbf{Implementation Details:} 
Consistent with NeRF~\cite{mildenhall2020nerf}, we optimize our model for each scene separately. We set $\lambda_{{\rm SN}}=1$, $\lambda_{\rm SL}=0.04$, $\lambda_{\rm SH}=0.1$, $\lambda_{\rm KP}=2$, and $\lambda_{\rm ED}=0.4$ via cross-validation. We use the Adam Optimizer~\cite{2015Adam} with an initial learning rate of $5\times 10^{-4}$ and set $\beta_{1}=0.9, \beta_{2}=0.999$. We train our model for 200k iterations on each scene, taking about 9 hours on a single NVIDIA RTX 2080 Ti GPU.

\textbf{Evaluation Metrics:} We use mean Intersection-over-Union (mIoU) to evaluate the semantic segmentation and $\mathcal{L}_1$ error to measure the performance of other tasks.

\begin{table}[t]
    \centering
    \resizebox{\linewidth}{!}{
    \begin{tabular}{c|ccccc}
    \hline
    Scene & SL($\uparrow$) & SN($\downarrow$) & SH($\downarrow$) & KP($\downarrow$) & ED($\downarrow$) \\ \hline
Office\_3 & 0.9345     &      0.0355          &   0.0423        &  0.0038        &  0.0155    \\ 
Office\_4    &  0.9162       &      0.0383          &     0.0503      &     0.0035     &  0.0150    \\ 
Room\_0 &  0.9707     &      0.0323          &     0.0293      &  0.0039        &   0.0209   \\ 
Room\_1 &   0.8757      &     0.0520           &     0.0495      &    0.0038      &  0.0202   \\ \hline
Avg. (Ours) & \textbf{0.9243} & \textbf{0.0395} & \textbf{0.0429} & \textbf{0.0038} & \textbf{0.0179}
\\\hline 
Avg. (Heuristic) & 0.8580 & 0.0424 & 0.0451 & 0.0059 & 0.0457\\
Avg. (Hybrid) & 0.7360 & 0.0593 & 0.0673 & 0.0055 & 0.0406\\ \hline 
    \end{tabular}
    }
    \vspace{-9pt}
    \caption{Performance of SS-NeRF on individual scene properties. SL: Semantic Labels; SN: Surface Normal; SH: Shading; KP: Keypoint; ED: Edge. 
    SS-NeRF reaches high quantitative scores for all the tasks, and also outperforms the two baselines, indicating that it is able to render accurate scene properties similar to the ground-truth.}
    \label{tab:main}
    \vspace{-10pt}
\end{table}

\begin{table}
    \centering
    \resizebox{0.8 \linewidth}{!}{
    \begin{tabular}{c|cccc} 
    \hline
        Setting & SL($\uparrow$) & SH($\downarrow$) & KP($\downarrow$) & ED($\downarrow$) \\ \hline 
        $\mathbf{F}_\mathrm{dec}^\mathrm{v}$ & 0.9072 & \bf 0.0503 & \bf 0.0035 &  \bf  0.0150 \\ 
        
         $\mathbf{F}_\mathrm{dec}^\mathrm{nv}$ 
         &\bf 0.9162 & 0.0794 & 0.0038 & 0.0183 \\ \hline
    \end{tabular}
    }
    \vspace{-9pt}
    \caption{Ablation results on different modeling for the four scene properties on Office\_4 in Replica. The view input is critical for SH, KP, and ED, but is redundant for SL.}
    \label{tab:ablation}
    \vspace{-10pt}
\end{table}

\subsection{Performance on Tasks Beyond RGB}
\label{sec:withrgb}

\begin{figure*}[t]
    \centering
    \includegraphics[width =  \linewidth]{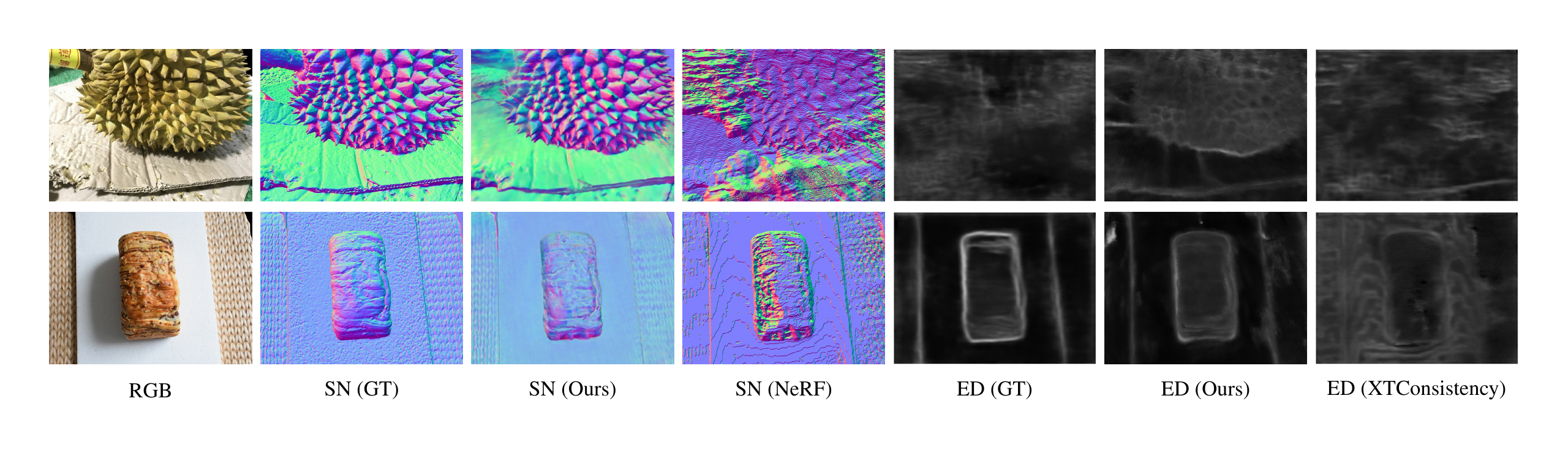}
    \vspace{-35 pt}
	\caption{Representative results on blendedMVS~\cite{yao2020blendedmvs}. SN (NeRF) is the normal derived from NeRF's depth; ED (XTConsistency) is the edge predicted from a well-trained model taking NeRF's normal as input. Our model outperforms both methods, indicating the capability and robustness of SS-NeRF.}
		\label{fig:blended}
		\vspace{-15 pt}
	\end{figure*}

We first build SS-NeRF for each individual scene property and evaluate them on Replica. We report the quantitative results in Table~\ref{tab:main}. Note that the main objective of this paper is to show that, with SS-NeRF, it is able to {\em synthesize} different scene properties paired with the rendered images; therefore, there is no existing work as baselines for a more comprehensive comparison. While there has been a large body of work on training discriminative models to predict the scene properties for {\em real} images, it is difficult to make an apple-to-apple comparison between these discriminative models and our synthesis model. Conceptually, the synthesis models can in principle produce infinite paired annotations, while the discriminative models are constrained by the given data.

However, to have a better understanding, we compare the model performance with one heuristic baseline and one hybrid baseline. \textbf{Heuristic Baseline (Heuristic)} estimates the annotations of the test view by finding the nearest view in the training set, and then mapping the source labels directly to the target view with perspective projection. \textbf{Hybrid Baseline (Hybrid)} trains a synthesis model (NeRF) and task-specific discriminative models separately. For novel test views, we first generate the color image corresponding to that pose, and then predict the annotations with the well-trained annotators. We adopt standard Taskonomy encoding-decoding architectures~\cite{zamir2018taskonomy} for Hybrid. We report the averaged results for all the scenes in Table~\ref{tab:main}.

From Table~\ref{tab:main}, we have the following observations: (1) SS-NeRF reaches a high performance for all five tasks, indicating that our model can well capture the original data distribution of the ground-truth; (2) SS-NeRF outperforms the heuristic baseline for all tasks, which verifies the accurate label quality generated by our SS-NeRF model; 
(3) SS-NeRF also outperforms the hybrid baseline for all the tasks, showing that it is non-trivial to synthesize paired color images and other scene properties, and that the shared 3D geometry and scene representation are critical for synthesizing different scene properties. 

We also visualize our rendered scene properties and compare with the corresponding ground-truth in Fig.~\ref{fig:main}. All of the images manifest the good quality of our SS-NeRF's novel view synthesis results for additional scene properties beyond RGB. Moreover, we have also conducted experiments on the real-world BlendedMVS dataset~\cite{yao2020blendedmvs} to verify the robustness. Two samples for SN and ED are shown in Fig.~\ref{fig:blended}. For SN, we compare with the normal derived by the NeRF's depth; and for ED, we compare with a even strong hybrid baseline, XTConsistency~\cite{zamir2020robust} that contains a powerful backbone and is pre-trained on the Taskonomy~\cite{zamir2018taskonomy} dataset, working on the NeRF's synthesized images.
Our model has obviously better visualizations for the challenging ``durian'' scenario on both tasks; For the simple ``bread'' scenario, we captures more details and our prediction for ED is closer to the ground-truth.

\subsection{Ablation Study}
\label{sec:ablation}

\noindent 
\textbf{Modeling with the two Decoders: }
In Sec.~\ref{sec:method}, we propose two branches for different scene properties. For each scene property except SN (special modeling), we choose the one with a better performance. In Table~\ref{tab:ablation}, we show the quantitative comparison between the two types of modeling for each scene property. We find $\mathbf{F}_\mathrm{dec}^\mathrm{v}$ works better for SH, KP, and ED, but cannot beat $\mathbf{F}_\mathrm{dec}^\mathrm{nv}$ for SL. This observation is consistent with the intuition: shading, keypoints, and edges vary from different viewing directions, but semantic labels keep the same. Therefore, the view input is critical for SH, KP, and ED but redundant for SL. This observation also indicates that SS-NeRF indeed learns a geometry-aware representation for the scenes.

\noindent 
\textbf{Modeling for RGB:}
The RGB color is a fundamental scene property and can facilitate the learning of the other properties. Here we ablate the key role of RGB color in two sets of experiments. First, we measure the averaged quality of the synthesized RGB images with peak signal-to-noise ratio (PSNR) for the basic NeRF model and all the variants of our SS-NeRF. The left part of Table~\ref{tab:rgb} shows that joint training of RGB and other scene properties will not affect the visual quality of the synthesized images. Including some properties, \eg, SL or SN, even improves the PSNR of the basic NeRF. Next, we build another variant of SS-NeRF for each scene property that removes the RGB color output (w/o RGB). The averaged performance among all the scenes is shown in Table~\ref{tab:rgb} right. Based on the result, we find that RGB supervision is crucial for understanding the scenes and learning other visual properties.

\begin{table}[t]
\begin{minipage}{0.33 \linewidth}
\centering
    \resizebox{\linewidth}{!}{
    \begin{tabular}{c|c}
   \hline
        Model & PSNR ($\uparrow$) \\\hline
        NeRF & 29.6916 \\
        SS-NeRF-SL & \bf 30.3127 \\
        SS-NeRF-SN & 30.0010 \\
        SS-NeRF-SH & 27.5024 \\
        SS-NeRF-KP & 29.7243 \\
        SS-NeRF-ED & 28.8324\\ \hline
    \end{tabular}
}    
\end{minipage}
\hfill
\begin{minipage}{0.63 \linewidth}
\centering
    \resizebox{\linewidth}{!}{
    \begin{tabular}{c|cc}
   \hline
   Property & w/o RGB (Avg.) &  w/ RGB (Avg.) \\ \hline
        SL ($\uparrow$) & 0.4235 & \bf 0.9209 \\
        SN ($\downarrow$) &  0.0433 & \bf 0.0409 \\
        SH ($\downarrow$) & 0.0548 & \bf 0.0420 \\
        KP ($\downarrow$) & 0.0437 & \bf 0.0037 \\
        ED ($\downarrow$) & 0.1659 & \bf 0.0187\\ \hline
    \end{tabular}
}    
\end{minipage}    
\vspace{-10pt}
\caption{Ablation study on the RGB color branch. \textbf{Left:} averaged PSNR measurement for the basic NeRF and SS-NeRF variants. Other scene properties will not affect the visual quality of the synthesized images. \textbf{Right:} performance comparison for the model with or without the RGB branch. RGB supervision is crucial for understanding the scenes and learning other visual properties.}
\label{tab:rgb}
\vspace{-8pt}
\end{table}

\subsection{Further Explorations within SS-NeRF}
\label{sec:applications}

\begin{table}[t]
\centering
\resizebox{\linewidth}{!}{
    \begin{tabular}{c|cccc} 
    \hline
    Setting & Office\_3 & Office\_4 & Room\_0 & Room\_1 \\ \hline 
         SH & 0.0423  &  0.0503 & 0.0293  &  0.0495 \\ 
         SH + SL & 0.0417(+) & 0.0479(+) & 0.0295(-) & 0.0432(+) \\
         SH + SN  & 0.0403(+) & 0.0471(+) & 0.0303(-) & 0.0445(+) \\
         SH + KP  & 0.0427(-) & 0.0478(+) & 0.0296(-) & 0.0473(+) \\
         SH + ED  & 0.0422(+) & 0.0483(+) & 0.0311(-) & 0.0501(-) \\
         SH + All  & 0.0415(+) & 0.0481(+) & 0.0318(-) & 0.0452(+) \\ \hline
    \end{tabular}
    }
    \vspace{-9pt}
    \caption{Model performance with additional tasks for shading. (+) indicates performance increasing, (-) indicates performance drop. SL consistently benefits the target SH task for nearly all the scenes, indicating a closer relationship between these two tasks.}
    \label{tab:relationship_on_seg}
    \vspace{-10pt}
\end{table}

\begin{table}

\centering
\resizebox{ \linewidth}{!}{
    \begin{tabular}{c|cccc} 
    \hline
       Settings & Office\_3 & Office\_4 & Room\_0 & Room\_1 \\ \hline
       Limited Views & 0.1171 & 0.0993 & 0.0685 & 0.1246 \\ \hline
       SL $\rightarrow$ SH & 0.0915 & 0.0886 & 0.0606 & 0.0982\\
       SN $\rightarrow$ SH & 0.0917 & 0.0911 & 0.0606 & 0.1002\\
       KP $\rightarrow$ SH & 0.0893 & 0.0864 & 0.0607 & 0.1016\\
       ED $\rightarrow$ SH & 0.0920 & 0.0864 & 0.0585 & 0.0965\\
       \hline
    \end{tabular}
    }
    \vspace{-9pt}
    \caption{Model performance with transfer learning. With the learned shareable knowledge from other scene properties, the transferred model can consistently have better performance, indicating the good generalization of SS-NeRF.}
    \vspace{-10 pt}
    \label{tab:transfer_learning}
\end{table}

\noindent
\textbf{Multi-task Learning:}
We instantiate SS-NeRF for each single scene property but it is able to simultaneously learn scene representations and shared knowledge within multiple visual tasks, so as to further benefit individual tasks. Taking SH as an example, we further build five variants under different task settings to conduct multi-task learning and also investigate whether other tasks can benefit from semantic segmentation in the framework of SS-NeRF. We first introduce the other four properties to be jointly trained with SH (denoted as SH + ``additional property''), and also build a variant that is trained for all the five properties (SH + All). 

We show the results in Table~\ref{tab:relationship_on_seg}. We have the following observations regarding the results: (1) SL consistently benefits the target SH task for all scenes except ``Room\_0,'' but the gap is marginal, indicating a closer relationship between the two tasks. It may be because the semantics label implicitly contains the texture and geometry information of the scene, which makes the model better estimate the shading. (2) Jointly training with all the tasks outperforms the single task model in three out of four scenes, indicating the general benefit of the knowledge from other scene properties. (3) Model performance also varies in different scenes, indicating that the task relationships might also rely on the scene structures, and the relationship among tasks might not be stationary for generative models. Interestingly, these observations are also consistent with those for discriminative models~\cite{zamir2018taskonomy,standley2019tasks}.

\noindent
\textbf{Knowledge Transfer:}
In additional to investigating multi-task learning, we explore the generalization of the learned scene representations by conducting transfer learning. Still taking SH as the target scene property, we first train our model with another source property and transfer the knowledge learned by the source to the target SH through initializing the learned encoding network $\mathbf{F}_\mathrm{enc}$. Different from previous experiments, here we focus on the typical transfer learning setting with limited data (6 training views) for the target property. The results are shown in Table~\ref{tab:transfer_learning}, for which ``Limited Views'' is the baseline without knowing any prior knowledge. We can find that with the learned shareable knowledge from other scene properties, the transferred model can consistently achieve better performance, indicating the effective generalization of the SS-NeRF framework.

\begin{table}[t]
\centering
\resizebox{\linewidth}{!}{
    \begin{tabular}{c|ccccc}
    \hline
    Data Setting & SL($\uparrow$) & SN($\downarrow$) & SH($\downarrow$) & KP($\downarrow$) & ED($\downarrow$) \\ \hline
GT &  0.5805            &        0.0394       &     0.0610      & 0.0051         &   0.0229  \\
SS-NeRF  &     0.5575     &     0.0434           &     0.0594      &         0.0048 &    0.0268 \\
GT + SS-NeRF &          \textbf{0.6178}    &       0.0394         &     0.0552      &    0.0048     &  0.0224 \\
GT + SS-NeRF-N &0.5929 & \textbf{0.0390} &\textbf{0.0531} &\textbf{0.0041} &\textbf{0.0206}
\\\hline 
    \end{tabular}
    }
\vspace{-10pt}
\caption{Comparison of the four data settings. GT: paired ground-truth data; SS-NeRF: paired synthesized data; GT+SS-NeRF: GT data and augmented data rendered by SS-NeRF (same pose); GT+SS-NeRF-N: GT data and augmented data rendered by SS-NeRF (novel pose). SS-NeRF synthesizes both visually realistic and useful data, so it can be used as an effective way of data augmentation to benefit the learning of other visual tasks.
}
\vspace{-15 pt}
\label{tab:augment}
\end{table}

\noindent
\textbf{Data Augmentation for Multi-task Learning:}
Given that we can render photo-realistic images and their corresponding scene property annotations, one natural, interesting question arises: How can we make use of these paired synthesized data? Inspired by~\cite{bao2021generative} and~\cite{devaranjan2020meta}, we design the following experiment. We adopt a task network (\ie, a standard discriminative model) to evaluate each task, and we train this model with four different data settings: (1) ground-truth (GT); 
(2) Paired RGB image and the corresponding annotations generated by SS-NeRF (SS-NeRF); 
(3) ground-truth and the augmented data generated by our model (GT+SS-NeRF);
(4) ground-truth and the augmented novel-view data synthesized by SS-NeRF (GT+SS-NeRF-N)
For the GT+SS-NeRF data setting, we generate paired data with the same poses as GT; and for the last setting, we generate data from novel views (averaged view from adjunct views in the training set).
For the task network, we adopt a standard Taskonomy encoding-decoding architecture~\cite{zamir2018taskonomy}. Different from the main experiment, we combine all the data from the four scenes together for evaluation. We train all the models for 200 epochs. The results are shown in Table~\ref{tab:augment}. We find: {\bf a.} GT and SS-NeRF have comparable performance, and SS-NeRF even outperforms GT in SH and KP, indicating the good quality of the data generated by SS-NeRF; {\bf b.} For all the five scene properties, including the augmented data, even from the same pose, can bring additional improvements; {\bf c.} and this improvement even gets higher for most tasks when we use augmented data from novel views. These results indicate that SS-NeRF can generate both visually realistic and useful data, making it attractive to be applied to benefit the learning of visual perception tasks. 


\begin{figure*}[t]
\vspace{-15 pt}
		\centering
        \includegraphics[width =  \linewidth]{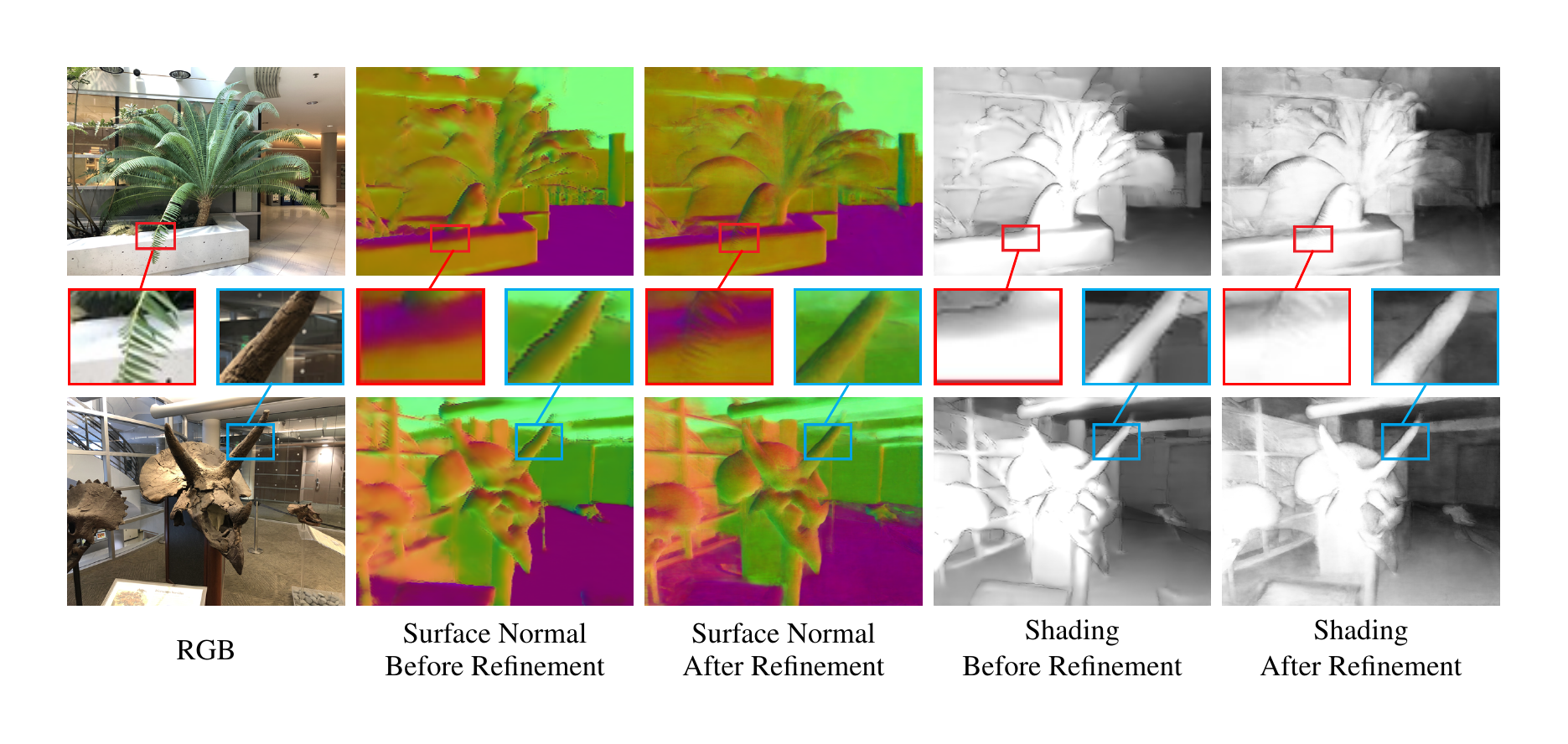}	
        \vspace{-41 pt}
		\caption{Surface normal and shading predictions with real-world images from the LLFF dataset. We use pre-trained annotators to obtain the initial labels which are noisy and flawed, and we retrain SS-NeRF with these labels. SS-NeRF can refine these flawed annotations and restore more details by joint modeling and understanding of the scenes.}
		\vspace{-15 pt}
		\label{fig:refinement}
\end{figure*}

\noindent
\textbf{Auto-Labelling for Real-World Scenes:}
One important application for the multi-task discriminative models is that they work as auto-labeller to annotate the real-world data, after pre-trained on synthetic or academic small-scale datasets. Our SS-NeRF model can be used as auto-labeller as well. Note that, different from discriminative models that directly operate on real images, our SS-NeRF simultaneously renders images and their per-pixel scene property annotations. Considering this difference, we introduce a {\em two-stage} procedure for leveraging SS-NeRF as auto-labeller. With some pre-trained discriminative models, we first produce initial ground-truth annotations. Such annotations are not guaranteed to be correct and could be even flawed -- \eg, they might be inconsistent across different views. Then, we train our SS-NeRF with these weak annotations. Because SS-NeRF can implicitly learn the 3D geometry and scene representation, it can {\em correct these inconsistencies} during network optimization. 
This refinement as auto-labeller is reminiscent of a denoising task~\cite{Zhi:etal:ICCV2021}, which  aims to correct the minor noisy ground-truth by learning from the majority of the accurate labels. However, the auto-labelling task is more challenging, since there is no guarantee that the majority of the annotations is accurate and the model has to detect and refine the correct labels based on the underlying 3D geometry.

Based on this insight, we move to a real-world dataset without annotations -- the LLFF dataset~\cite{mildenhall2019local}. We also use a pre-trained annotator~\cite{zamir2020robust} to generate weak annotations for this dataset (2nd and 4th columns in Fig.~\ref{fig:refinement}). Due to the data distribution gap between LLFF and the Taskonomy dataset, the quality of these annotations is quite poor, \eg, for the surface normal, there are sharp faults in the object boundaries. Then we train SS-NeRF with these flawed annotations and we show the results for surface normal and shading on two scenes of the LLFF dataset in Fig.~\ref{fig:refinement}. It is clear to see that our SS-NeRF produces smoother results, contains more details, and reflects better 3D structures of the scene. We argue that the refinement comes from the joint modeling and understanding of the scenes, inherent within the SS-NeRF framework, showing the capability of our model in scene analysis. In addition, this general idea of auto-labelling and refinement can be in principle applied to other real-world data and jointly work with other discriminative models.

\vspace{-3 pt}
\subsection{Limitations and Future Work}
\label{sec:limitation}
There are two major limitations for our SS-NeRF model: (1) SS-NeRF builds upon the original NeRF model, which is scene-dependent, making it hard to transfer the learned knowledge from one scene to another; (2) SS-NeRF requires accurate and dense pose annotations to learn scene representations, which might not be accessible for all the datasets (\eg, Taskonomy~\cite{zamir2018taskonomy}). Notice that these limitations are essentially from the original NeRF model and some following work has provided promising solutions~\cite{schwarz2020graf,gu2021stylenerf,yu2021pixelnerf,attal2021t,wang2021nerfmm}. Similar techniques can be introduced to our SS-NeRF framework to further enhance the model capability. 

Our work provides a first versatile representation for scene property synthesis based on neural radiance fields. The high-level motivation is that the underlying 3D geometry and scene representation from NeRF enable the knowledge sharing across different tasks, therefore make it able to extend from color image synthesis to other scene properties. Investigating similar strategies for other formats of scene representations, such as point clouds~\cite{wiles2020synsin} and meshes~\cite{hu2021worldsheet}, can also be promising directions for future research.

\section{Conclusion}
\label{sec:conclusion}
This work shows that a comprehensive scene representation with implicitly encoded 3D geometry and scene structure, powered by the NeRF-style architecture, can be useful for not only RGB image synthesis tasks, but also various visual tasks. Inspired by this, we propose a unified framework SS-NeRF that allows knowledge and representation sharing across different tasks. This novel strategy of solving visual perception problems with a synthesis model provides a different perspective for multi-task learning, which is normally tackled in the context of discriminative models. We further show some interesting observations and promising applications within this synthesis model.

{\small
\bibliographystyle{ieee_fullname}
\bibliography{egbib}
}

\end{document}